# Highlights

## Transfer Language Selection for Zero-Shot Cross-Lingual Abusive Language Detection

Juuso Eronen,Michal Ptaszynski,Fumito Masui,Masaki Arata,Gniewosz Leliwa,Michal Wroczynski

- Zero-shot cross-lingual transfer can yield good results for abusive language detection
- Linguistic similarity metrics can be used to find an optimal cross-lingual transfer language, at least for abusive language detection
- Choosing a transfer language by intuition, for example by purely looking at the same language family, is not optimal
- The World Atlas of Language Structures can be quantified into an effective linguistic similarity metric

# Transfer Language Selection for Zero-Shot Cross-Lingual Abusive Language Detection

Juuso Eronen[a,∗], Michal Ptaszynski[a], Fumito Masui[a], Masaki Arata[a], Gniewosz Leliwa[b] and Michal Wroczynski[b]

[a]*Kitami Institute of Technology, 165, Koencho, Kitami, 090-0015, Hokkaido, Japan*
[b]*Samurai Labs, Aleja Zwycięstwa 96/98, Gdynia, 81-451, Poland*

ABSTRACT

We study the selection of transfer languages for automatic abusive language detection. Instead of preparing a dataset for every language, we demonstrate the effectiveness of cross-lingual transfer learning for zero-shot abusive language detection. This way we can use existing data from higher-resource languages to build better detection systems for low-resource languages. Our datasets are from seven different languages from three language families. We measure the distance between the languages using several language similarity measures, especially by quantifying the World Atlas of Language Structures. We show that there is a correlation between linguistic similarity and classifier performance. This discovery allows us to choose an optimal transfer language for zero shot abusive language detection.

## 1. Introduction

Harmful language in online communication can cause serious consequences to its victims. In the worst cases, it can lead to self-mutilation or suicide, or, on the contrary, to a retaliation assault on their perpetrators [1]. There have been multiple attempts to automate the detection of offensive content online [2, 3, 4] in order to reduce the human effort needed in prevention of the uncontrolled spread of harmful content on social media. Even though there are thousands of languages used in different social media platforms, the research on the detection of harmful content has only been done with a handful of them, mostly in English [5, 6], Japanese [7, 8] and Polish [9].

Being able to detect harmful language like hate speech and cyberbullying (CB) also in low-resource languages would be a great aid, because social media is used in thousands of languages, of which only a small fraction have proper data to train the detection models on. It is also important to detect offensive content as urgently and effectively as possible because of its increasing prevalence and serious consequences [10]. Users' realization of the anonymity of online communications is one of the factors that make this activity attractive for harassers and bullies since they rarely face consequences of their improper behavior. The problem was further exacerbated by the popularization of smartphones and tablet computers that enable almost continuous usage of social network services (SNS) anywhere, at home, work/school or in motion [11].

Messages that can be identified as abusive usually encourage violence against a person or group based on race, religion or sexual orientation, etc. (hate speech) [12] or ridicule someone's personality, body type or appearance, or include slandering or spreading rumors about the individual (cyberbullying). This may drive its victims to even as far as self-mutilation or suicide, or, on the contrary, to a retaliation assault on their perpetrators [1]. The offenses are carried out by exploiting open online means of communication, such as Internet forum boards, or SNS to convey harmful and disturbing information about private individuals, often children and students [13].

In certain countries, such as in Japan, the issue has been severe enough to be seen at ministerial level [14]. As one of the ways to solve the issue, Internet Patrol (IP) consisting of school workers has begun to track online forum pages and SNS featuring cyberbullying content. Unfortunately, as IP is carried out manually, reading through vast numbers of websites and SNS content makes it an uphill battle. To aid in this struggle, some research have started to develop methods for automatic detection of CB [7, 8, 6, 15].

∗Corresponding author
eronen.juuso@gmail.com (J. Eronen)
ORCID(s): 0000-0001-9841-3652 (J. Eronen)

Recently, the research of automatic hate speech detection has expanded to dealing with low-resource languages. This has come with new challenges as these languages lack proper datasets to be used for training the detection models. To get around this problem, it has been shown that with cross-lingual transfer, the performance on low-resource languages can be improved by leveraging knowledge from other higher resource languages. This has also been demonstrated to be an effective technique in improving offensive content detection in low resource languages by using cross-lingual word embeddings and multilingual transformer models [16, 17, 18, 19].

However, choosing the optimal language for the transfer remains widely an understudied problem. Usually, it is up to the individual (researcher, or ML practitioner) to decide experimentally or by pure intuition which language might be suitable for the transfer, based on their field experience and accumulated theoretical knowledge. For example, one could select the transfer language by looking at languages belonging to the same language family as the target language [20]. But this does not necessarily mean that the two languages would share the same linguistic features [21].

In order to contribute to further understand and solve this problem, we propose a method for selecting transfer languages for automatic hate speech detection. We show that there is a correlation between linguistic similarity and classifier performance, which allows us to choose the optimal transfer language using different linguistic similarity metrics. We also show that it is possible to achieve good performance on the target language by training only on the source language using multilingual transformer models. To select the optimal source language for transfer, we propose to quantify the features of languages to compute a metric that can be used in comparing the closeness of languages using their linguistic properties.

There are some existing metrics that can be used to estimate the similarity between languages by using linguistic features [22, 23, 24]. However these metrics only take a single feature or a handful of features into account. To fill this gap, we propose a new linguistic similarity metric based on the World Atlas of Language Structures (WALS) [25], which contains almost two hundred different features. This way, we can better quantify all aspects of the languages and not simply rely on one, or a few features only.

We used datasets from seven different languages, namely English, German, Danish, Polish, Russian, Japanese and Korean. The languages were chosen as there were high quality datasets for those languages and because the languages represent different language families (English, German, Danish - Germanic; Polish, Russian - Slavic; Japanese, Korean - Koreano-Japonic language family). This allows us to study the efficacy of transfer learning between and within language family groups. The datasets mainly contained between 3,000 to 35,000 samples, with the exception of the Korean dataset, which was substantially larger with almost 200,000 samples.

We hypothesize that the transfer learning performance correlates with the similarity of the source and target languages. To confirm that we used multilingual transformer models, namely Multilingual BERT [26] and XLM-RoBERTa [27] in our experiments. We fine tuned the models by using each of the languages as source and target and calculated the linguistic similarity between the languages using three metrics, EzGlot, eLinguistics and a quantified model based of WALS. Then we checked the correlation between the performance and linguistic similarity to show the effectiveness of our method.

The paper outline is as follows. In Section 2 we describe previous research in all areas that are addressed in this paper. In Section 3 we describe all the datasets applied to this research and present their features. In Section 4 we go through the classification methods and the linguistic similarity metrics used in this research. In Section 5 we go through all the results from the conducted experiments. In Section 6 we discuss the results in general and bring out the most interesting findings in relation to the research goals.

## 2. Previous Research
### 2.1. Abusive Language Detection

Even though the issue of hate speech and cyberbullying has been researched in social sciences and psychology for over fifteen years [13, 28], the first attempt to use information technology to help solve the problem was done by Ptaszynski et al. [7, 8] in 2010 who performed affect analysis on a small CB dataset and discovered that the use of vulgar words were the distinctive features for CB. They trained an SVM classifier using a lexicon of such words and with multiple optimizations, they managed to detect CB with an F-score of 88.2%. However, as the amount of data increased, it caused a decrease in results, which caused the authors to abandon SVM as not ideal for language ambiguities typical for CB.

In other research, Sood et al. [29] focused on detecting personal insults and negative influence which could at most cause the Internet community to fall into recession, meaning if the harmful content would be left uncontrolled, people would start to leave the community. Their study used single words and bigrams as features, and weighted them using Boolean weighting (1/0), term frequency and tf-idf. These were used to train an SVM classifier. Their dataset was a corpus collected from multiple online fora, totaling at six thousand entries. They used a crowd-sourcing approach (Mechanical Turk) with non-professional laypersons hired for the classification task to annotate the data.

Later, Dinakar et al. [6] introduced their method to detect and mitigate cyberbullying. Their paper had a wider perspective, as they did not focus only on the detection of cyberbullying, but also included methods for mitigating the problem. This was an improvement compared to previous research. Their classifiers scored up to 58-77% of F-score on an English dataset. The results varied depending on the type of harassment they were attempting to classify. The best classifier they proposed was again an SVM, which further confirms the effectiveness of SVMs for detecting cyberbullying, similarly to the research done in 2010 using a Japanese dataset [7].

An interesting work was done by Kontostathis et al. [5], who performed a thorough analysis of bullying entries on Formspring.me. They identified usual bullying patterns and used a machine learning method based on Essential Dimensions of Latent Semantic Indexing (EDLSI) to apply them in classification. In a different research, Cano et al. [30] introduced a Violence Detection Model (VDM), a weakly supervised Bayesian model. This simplified the problem and made it more feasible for untrained annotators to work with. The datasets were extracted from violence-related topics on Twitter and DBPedia.

Nitta et al. [31] proposed a method extending Turney's SO-PMI-IR score [32] to automatically detect harmful entries. The seed words were categorized as abusive, violent and obscene. Initially, they obtained a very high over 90% precision, however, a re-evaluation of their method two years later unfortunately showed a great decrease in performance over the span of two years [15]. They hypothesized that this could be the cause of external factors like Web page re-ranking, or changes in SNS user policies, etc. The method was improved by acquiring and filtering new harmful seed words automatically with some success, but they were unable to achieve results close to the original performance. Later an automatic method for the seed word acquisition [33] was developed with positive results. However, this method was deemed inefficient compared to a more direct machine learning based method.

The offensive language detection method by Sarna et al. [2] was based on features like "bad words", positive/negative sentiment words, etc., to estimate user credibility. These features were applied to commonly used classifiers like Naive Bayes and SVM. The obtained classification results were further used in User Behavior Analysis model (BAU), and User Credibility Analysis (CAU) model. Even though their approach included the use of phenomena such as irony, or rumors, in practice they unfortunately only focused on messages containing "bad words." Moreover, neither the words themselves, the dataset, nor its annotation schema were sufficiently described in the paper.

Ptaszynski et al. [34] suggested a pattern-based language modeling system. Identified as ordered combinations of sentence elements, the patterns were extracted with the use of a Brute-Force search algorithm. They reported promising initial findings and further developed the system by adding several data pre-processing techniques [9]. In 2017, Ptaszynski et al. [35] proposed a method of using Linguistically-backed preprocessing methods and implemented the notion of Feature Density to find an optimal way to preprocess the data in order to achieve higher performance, particularly with Convolutional Neural Networks. The experiments performed on actual cyberbullying data showed a major advantage of this approach to all previous methods, including the best performing method so far based on Brute Force Search algorithm. The method was later confirmed to also be useful in finding the best feature sets to be used in training to reduce the redundant experiment runs [36].

Vidgen and Derczynski [4] examined over sixty hate speech datasets in 2020. They provided insights into the contents of the datasets, their annotation and the formulation of the associated tasks. They also announced hatespeechdata.com [1], a repository for online abusive content training datasets, in order to make quality data more accessible. Lastly, they provided outlines on best practices for the creation of datasets for online abuse detection.

In the recent years the research in offensive language detection has gained more popularity and has mainly focused on using recurrent neural networks and pretrained language models [37, 38, 39, 40]. Also, the popularization of multilingual neural models has made it possible to train models for low-resource languages by utilizing transfer learning. As with cross-lingual transfer, the performance on low-resource languages can be improved by leveraging knowledge from other higher resource languages [41].

Ranasinghe et al. [16, 17] showed the effectiveness of cross-lingual transfer in offensive language identification in Hindi, Spanish, Danish, Greek and Bengali. Their work showed that multilingual transformer models like mBert and XLM-R can use the knowledge gained from higher resource languages to gain an improved performance on a low-resource target. Also, the models scored comparatively high without any data from the target language, demonstrating the power of cross-lingual pre-training.

---

[1]http://hatespeechdata.com

Similar results were obtained by Bigoulaeva et al. [18] with English and German. They also discovered that using unlabeled samples from the target language can be used to increase performance. Finally, Gaikwad et al. [19] noticed that transfer learning from Hindi outperformed other languages when classifying entries in Marathi, suggesting a relation between cross-lingual transfer performance and language similarity.

## 2.2. Measuring Linguistic Similarity

The relation between the difficulty of language learning and language similarity was already discussed in a book by Ringbom [42] in 2006. He presented an example about the Finnish language scene as a demonstration of the importance of cross-linguistic similarity in foreign language learning [43]. In short, he showed that Swedish-speaking Finns have a greater advantage in learning English than Finnish-speaking Finns due to the closer relation between Swedish and English.

Cottorell et al. [44] showed that not all languages are equally difficult to model. They showed that a correlation exists between the morphological richness of a language and the performance of the model, meaning that more complex languages are more difficult to model. This gives a hint that simple languages might not work as well if used as the cross-lingual transfer source for more complex languages. This also shows that relatedness of languages should not be the only criteria when choosing the transfer source and that other features of the languages should also be considered.

There have been some attempts in quantifying a similarity metric for languages from different linguistic features. However, most of these metrics rely on only a single or a few aspects of languages. For example, one can calculate a genetic proximity score between two languages by comparing the consonants contained in a predefined set of words [23], while taking into account the order in which these consonants appear in the words. This gives information about the direct relatedness of the compared languages. However, we noticed that as used languages become more distant, there is a significant increase in errors as more and more accidental similarities in consonants are introduced. While being easy to calculate, this method completely disregards other aspects of languages like semantic, morphological, and syntactic similarity.

Another way to compute a similarity metric is to consider how similar the vocabularies of two languages are. Lexical similarity is used as the basis of some similarity metrics like EzGlot [24]. This metric is calculated using lexical similarity between the two compared languages while also considering the number of words these languages share with other languages. This way, it is possible to calculate the similarity between the two languages in relation to similarities to all other languages.

There has been some research attempting to compute a similarity metric for languages using multiple aspects of languages. Aggarwal et al. [22] proposed STL, a metric based on Semantic, Terminological (lexical) and Linguistic (syntactic) similarity of languages. The method outperformed previous similarity metrics that were using only a single one of these aspects [45, 46]. They noticed that the terminological measures showed a significant contribution compared to the two other aspects. However, the structure of the used vocabulary dataset needs to be in the form of an ontology. This and the lack of the available languages for the used dataset made it not possible to be utilized in this research.

## 2.3. Transfer Language Selection

Choosing the optimal language for the transfer remains widely an unanswered problem. Usually, it is up to the practitioner's consideration to decide which language to use. This is usually done experimentally or simply by intuition [47]. For example, Cottorell and Heigold [20] focused on using languages belonging to the same language family as the target language for a more successful transfer. However, using languages from the same language family does not guarantee them sharing the same linguistic features and the languages could be distant for example when looking at the complexity of grammar [21].

Multiple research [48, 49, 50] discovered that using multiple high-resource languages at the same time for transfer yielded higher results than selecting only a single language. However, their methods do not consider the actual relation between the source and the target languages and the amount of contribution of each of the languages to the total score. Nooralahzadeh et al. [51] also discovered that languages which share certain morphosyntactic features tend to work better for cross-lingual transfer.

The discovery by Gaikwad et al. [19] suggests a relation between cross-lingual transfer performance and language similarity. They used multiple languages, specifically Bengali, Greek, English, Turkish and Hindi as cross-lingual transfer sources for classifying entries in Marathi language. Out of all of the used languages, Hindi had the highest performance as the transfer source, while also being the language with the closest relation to Marathi out of the used languages. This discovery hints a solution to the problem of cross-lingual transfer language selection, at least for offensive language detection.

**Table 1**
Statistics of the applied datasets

|  | **English** | **German** | **Danish** | **Polish** | **Russian** | **Japanese** | **Korean** |
| --- | ---: | ---: | ---: | ---: | ---: | ---: | ---: |
| Category | CB | Offense | Offense | Offense | Toxic | CB | Hate |
| Number of samples | 12,772 | 8,407 | 3,289 | 34,953 | 14,412 | 4,096 | 189,995 |
| Number of offensive samples | 913 | 2,838 | 425 | 7,367 | 4,826 | 2,048 | 89,999 |
| Number of non-offensive samples | 11,859 | 5,569 | 2,864 | 27,586 | 9,586 | 2,048 | 99,996 |
| Avg. length (chars) of a sample | 125.5 | 136.4 | 102.4 | 165.4 | 176.5 | 39.3 | 50.2 |
| Avg. length (words) of a sample | 28.7 | 19.0 | 18.8 | 26,0 | 27.9 | 14.9 | 19.9 |
| Avg. length (chars) of an off. sample | 115.4 | 131.9 | 143.3 | 89.8 | 141.4 | 38.3 | 55.9 |
| Avg. length (words) of an off. sample | 26.8 | 18.5 | 26.8 | 14.2 | 22.4 | 14.5 | 22.1 |
| Avg. length (chars) of a safe sample | 126.3 | 138.7 | 96.3 | 185.7 | 194.2 | 40.2 | 45.0 |
| Avg. length (words) of a safe sample | 26.8 | 19.3 | 17.7 | 29.2 | 30.7 | 15.3 | 17.9 |
| Split (train/eval) | 80/20 | 60/40 | 90/10 | 83/17 | 80/20 | 80/20 | 80/20 |

### 2.4. Contributions of This Study

This research aims to answer the need of developing a method of selecting languages for cross-lingual transfer learning. The current methods are mainly based on the individual's own judgement based on their field experience and accumulated theoretical knowledge or simply choosing languages from the same language family [20]. The problem with the current selection methods are that they are completely unoptimized and prone to bias from the practitioner. In fact, one could argue that there is no systematic method that would give an actual score or ranking for the transfer language candidates.

Our approach to this problem is to explore, whether different linguistic similarity metrics could be used for finding the optimal candidates for cross-lingual transfer. Supported by the findings of Gaikwad et al. [19], we hypothesize that linguistic similarity correlates with cross-lingual transfer efficacy, meaning that using more similar languages would yield a higher classification score. In practice, we fine tune cross-lingual pretrained language models, specifically mBERT and XLM-R, separately on each of our proposed languages (English, German, Danish, Polish, Russian, Japanese, Korean) and then perform zero-shot classification on the rest of the languages of the proposed set.

Another goal is trying to aid in tackling the issues of cyberbullying and hate speech, which, in addition to being serious social problems, have also become more prevalent and radical [52] in the recent years due to the popularization of online social media. This constantly calls for increasing needs for solving the problem. With a proper solution for transfer language selection, the detection performance on low-resource languages would be greatly improved. This is because social media is used in thousands of languages, of which only a small fraction have proper data to directly train the detection models on.

## 3. Datasets

In order to confirm our hypothesis, we used offensive language (hate speech, cyberbullying) datasets from seven different languages, namely English, German, Danish, Polish, Russian, Japanese and Korean. We chose these languages as they had high quality datasets compared to other options and because the languages represent three different language families (English, German, Danish - Germanic; Polish, Russian - Slavic; Japanese, Korean - Koreano-Japonic language family). This also allows us to study the efficacy of transfer learning between and within language family groups. Preliminarily, we had a few datasets of good quality (English, Polish, Japanese). We tried to keep the quality high also for the other languages to the best of our ability. For Germanic languages, there are some good quality datasets available, but then same cannot be said about Russian and Korean and we had to loosen our standards for these two languages. Key statistics of the applied datasets are shown in Table 1. Training and evaluation splits were retained from original datasets if possible, otherwise datasets were split to 80% training and 20% evaluation.

### 3.1. English Dataset

The first dataset for our experiments was the Kaggle Formspring Dataset for Cyberbullying Detection [53]. There was one major problem with the original dataset however, as the original annotations for the data were carried out by untrained laypeople. It has been proven before that the annotations for topics like online harassment and cyberbullying should be done by experts [3]. Therefore, the dataset was re-annotated with the help of experts with sufficient psychological background to assure high quality annotations [54]. In our research we applied the re-annotated version for more accurate results.

The dataset contains approximately 300 thousand of tokens. There was no visible difference in length between the posted questions and answers, both being approximately 12 words long on average. On the contrary, the harmful (CB) entries were usually slightly but insignificantly shorter compared to the non-harmful (non-CB) samples (approx. 23 vs. 25 words). The amount of harmful samples was also substantially smaller compared to the amount of non-harmful samples, around 7% of the whole dataset, which is approximately the same as the real-life amount of profanity encountered on SNS [3].

### 3.2. German Dataset

The German dataset originates from the 2018 GermEval offensive language identification shared task [55] and contains around 8,000 entries collected from Twitter. They decided against collecting a natural sample as it would have ended up making the portion of offensive tweets too small. They also decided against sampling by specific query terms. Instead they heuristically identified users that regularly post offensive tweets and sampled their timeline. This allowed for more offensive tweets in comparison to taking a natural sample without biasing the dataset with specific terms. However this caused certain topics to dominate in the extracted data, like the situation of migrants or the German government. So they decided to bias the data collection by sampling further arbitrary tweets containing common terms found in these topics like names of politicians and the word "refugee".

There are some rules regarding the selection of the tweets put up by the authors. Each tweet is written in German and contains at least five ordinary alphabetic tokens. Also the tweets do not contain any URLs and retweets were not allowed. In splitting the data into training and test sets, the authors decided to assign any given user's complete set of tweets to either the training set or the test set. This way, they could avoid the fact that the classifiers could benefit from learning user-specific information.

### 3.3. Danish Dataset

Sigurbergsson and Derczynski [56] collected the Danish dataset from Facebook and Reddit. The final dataset contains 3600 user-generated comments, 800 from Ekstra Bladet on Facebook, 1400 from r/DANMAG and 1400 from r/Denmark.

After collecting the initial corpus, they published a survey on Reddit in order to maximize the number of user-generated comments belonging to the classes of interest (offensive language), where they asked Danish speaking users to suggest offensive, sexist, and racist terms. This lexicon was then used to find potentially-offensive comments. A subset was then taken from the comments remaining in the corpus to fill the remainder of the final dataset. This helped to ensure that the data would have coverage beyond just the terms found in the lexicon.

They based the annotation procedure on the guidelines and schemas presented by Zampieri et al. [57]. As a warm-up procedure, the first 100 posts were annotated by two annotators and the results were compared. This exercise was used to refine the understanding of the task and to discuss the mismatches in these annotations. They assessed the similarity of the annotations using a Jaccard index.

### 3.4. Polish Dataset

The Polish corpus is a combination of two datasets. One originates from PolEval workshop from 2019 [58], collected from Twitter discussions. Another one was collected from Wykop[2], which is a Polish social networking service. As feature selection and feature engineering have been proven to be integral parts of cyberbullying detection [35, 59], the entries are provided as such, without additional preprocessing to allow researchers using the datasets apply their own preprocessing methods. The only preprocessing applied to the dataset was done only to mask private information, such as personal information of individuals (usernames, etc.).

The datasets were initially annotated by laypeople, and further corrected by experts in case of disagreements. The laypeople agreed on majority of the annotations. This is mostly due to the fact that the annotators mostly agreed upon non-harmful entries, which take up most of the dataset. When considering the harmful class, the annotators only fully agreed upon less than two percent of the entries. Moreover, some of the fully-agreed entries needed to be corrected to the opposite class in the end by the expert annotator, which shows that using laypeople does not provide accurate enough annotations in the field of offensive language identification. It could be said that layperson annotators can tell with a decent level of confidence that an entry is not harmful (even if it contains some vulgar words), and they can spot, to some extent, if the entry is somehow harmful. Though in most cases they are unable to provide a reasoning for their choice. This provides further proof that for specific problems such as cyberbullying, an expert annotation is required [3].

---

[2]https://www.wykop.pl

### 3.5. Russian Dataset

The Kaggle Russian Language Toxic Comments Dataset [3] is the collection of annotated comments from Russian online communication platforms. 2ch, which is a popular Russian anonymous image board and Pikabu, which could be considered the Russian equivalent of Reddit. The dataset was published on Kaggle in 2019. It consists of a total of 14,412 comments, out of which 4,826 texts are labeled as toxic, and the remaining 9,586 are labeled as non-toxic. The average length of the comments is around 175 characters. The minimum length being 21, and the maximum length being 7403 characters. The annotators of this dataset are unknown so unfortunately we cannot say anything about its quality. Although the annotations were validated with the help of Russian language speakers (laypeople) using a crowd sourcing application [60].

### 3.6. Japanese Dataset

The Japanese cyberbullying dataset we used for our experiments was created by combining two separate datasets. The first of which was originally described by Ptaszynski et al. [7], and also widely used in other research [31, 34, 9, 15, 35]. It contains 1,490 harmful and 1,508 non-harmful entries written in Japanese, collected from unofficial school websites and fora. The original data was provided by the Human Rights Research Institute Against All Forms for Discrimination and Racism in Mie Prefecture, Japan. The entries were collected and labeled by Internet Patrol members (expert annotators) with the help of the government supplied manual [14]. The instructions given by the manual are briefly described below.

The definition given by the Ministry of Education, Culture, Sports, Science and Technology (MEXT) of Japan suggests that cyberbullying occurs when a person is directly offended on the Internet. This includes publication of the person's identity, personal information and other aspects of privacy. Thus, as the first distinguishable features for cyberbullying, MEXT identifies private names (also initials and nicknames), names of organisations and affiliations and private information (address, phone numbers, personal information disclosure, etc.)

In addition, cyberbullying literature reveals vulgarities as one of the most distinguishing characteristics of cyberbullying [13, 61]. Also according to MEXT, vulgar language and cyberbullying can be distinguished from each other as cyberbullying conveys offenses against specific individuals. In the prepared dataset, all entries containing at least one of the above characteristics were listed as harmful.

The second Japanese dataset was collected from Twitter by Arata [62]. The dataset consists of random tweets that were collected during a one-week period in July of 2019. The collected information included the ID, the date and time of posting, the username, the text body, and the URL of the tweet. In addition, tweets written in other languages than Japanese, tweets submitted by bots and tweets consisting of less than five characters were filtered out from the collection.

The guidelines for the annotations were based on the information provided by Safer Internet Association [4] [5] and also on the research by Takenaka et al. [63]. The tweets were organized into seven different categories based on the above sources. The annotation work was carried out in two stages, primary annotation and secondary annotation. This was done in order to reduce the burden on each annotator and to not allow the work of each annotator influence one another. First, the annotators determined whether a tweet will fall into either harmful or not. After that, one of seven specific categories (illegal acts, prostitution, suicide, abuse/slander/bullying, obscenity, cruelty, other) was assigned to the tweet in the second annotation. In total, 30,425 tweets were annotated by six annotators who attended an Internet patrol course organized by the Hokkaido Police. For a single tweet, the primary annotation was done by a single annotator and the secondary annotation was carried out by two annotators. After this, all of the annotations were validated by at least three people.

---

[3] https://www.kaggle.com/blackmoon/russian-language-toxic-comments
[4] https://www.saferinternet.or.jp/wordpress/wp-content/uploads/bullying_guideline_v3.pdf
[5] https://www.safe-line.jp/wp-content/uploads/safeline_guidelines.pdf

The number of tweets annotated as harmful, defined in the guidelines mentioned above, was less than 2%. In addition, when looking at each category, about 75% of tweets were given the category of "abuse/slander/bullying". Furthermore, when validating the contents of tweets in this category, 202 tweets were additionally annotated as "prostitution", out of which almost half were duplicated. The 115 tweets that were not annotated in either of these categories were mostly related to "obscenity". In this study, a balanced random subset, consisting of 552 harmful and 546 non-harmful entries, of the 30,425 tweets was used.

### 3.7. Korean Dataset

The Kaggle Korean Hate Speech Dataset [6] is a collection of Korean hate speech text data. Composed of hateful and discriminatory comments scraped from Korean alt-right website (Daily Best). The dataset was published on Kaggle in 2020. It consists of almost 190,000 comments, where 90,000 texts were labeled as hate speech, and 100,000 were labeled as normal. The average length of comments is around 50 characters. The annotators of this dataset are unknown so we are unable to say anything regarding its quality.

## 4. Methods

### 4.1. Classification

In the experiment we applied the following classification algorithms. The assumption was that these multilingual transformer classifiers can generalize sufficiently well in a zero-shot cross-lingual setting [64]. This means using no data from the target language while fine-tuning only with data from one of the other proposed languages.

**Multilingual BERT** (mBERT) [26] is the multilingual version of BERT, which stands for Bidirectional Encoder Representations from Transformers. It is based on transformer [65], an attention mechanism that learns contextual relations between words (or sub-words) in a text. Transformer models also introduced a capability for the models to read text input in both directions at once, not only sequentially from left-to-right or right-to-left. With this bidirectional capability, BERT is pre-trained on two NLP tasks, Masked Language Modeling and Next Sentence Prediction. The objective of Masked Language Modeling is to hide a word in a sentence and have the algorithm predict what word has been hidden, or masked, based on the word's context. The objective of Next Sentence Prediction is to have the algorithm predict whether two of any given sentences have a connection, either logical or sequential, or whether their relationship is random.

Even though mBERT has not been trained on cross-lingual data, it has still showed some cross-lingual capabilities [66]. This includes achieving a good performance in various zero-shot transfer tasks, even outperforming the usage of some cross-lingual embeddings [67]. It is hypothesized that this generalization ability comes from having word pieces used in all languages (numbers, URLs, etc) which have to be mapped to a shared space forces the co-occurring pieces to also be mapped to a shared space, thus spreading the effect to other word pieces, until different languages are close to a shared space [68].

**XLM-RoBERTa** (XLM-R) [27] is a cross-lingual transformer model that is also trained on the Masked Language Model objective and is capable of processing text from 100 separate languages. XLM-R is trained on around 2.5tb of CommonCrawl data. The model's training routine is the same as the monolingual RoBERTa model[69], meaning, that the sole training objective is the Masked Language Model. There is no Next Sentence Prediction like in BERT.

XLM-R outperforms mBERT and XLM on a variety of cross-lingual benchmarks, which include zero-shot transfer tasks [70]. It also performs particularly well on low-resource languages. Interestingly, XLM-R is also very competitive when comparing to state-of-the-art monolingual models, which shows the possibility of multilingual modeling without losing performance in a monolingual setting [27].

### 4.2. Linguistic Similarity Metrics

In order to measure the correlation between cross-lingual transfer performance and linguistic similarity, we needed a language similarity metrics that could quantify the properties of the proposed languages. In this study we applied three different linguistic similarity metrics. One of these metrics is a novel metric we quantified from the linguistic features presented in WALS [25].

---
[6]https://www.kaggle.com/captainnemo9292/korean-hate-speech-dataset

**Table 2**
eLinguistics distance metric

|          | Danish | English | German | Japanese | Korean | Polish | Russian |
|----------|--------|---------|--------|----------|--------|--------|---------|
| Danish   | 0.0    | 20.6    | 38.2   | 95.2     | 97.2   | 68.2   | 66.2    |
| English  | 20.6   | 0.0     | 30.8   | 88.3     | 90.0   | 66.9   | 60.3    |
| German   | 38.2   | 30.8    | 0.0    | 87.4     | 95.5   | 68.1   | 64.5    |
| Japanese | 95.2   | 88.3    | 87.4   | 0.0      | 88.0   | 93.3   | 93.3    |
| Korean   | 97.2   | 90.0    | 95.5   | 88.0     | 0.0    | 89.5   | 89.5    |
| Polish   | 68.2   | 66.9    | 68.1   | 93.3     | 89.5   | 0.0    | 5.1     |
| Russian  | 66.2   | 60.3    | 64.5   | 93.3     | 89.5   | 5.1    | 0.0     |

**Table 3**
EzGlot similarity metric

|          | Danish | English | German | Japanese | Korean | Polish | Russian |
|----------|--------|---------|--------|----------|--------|--------|---------|
| Danish   | 100    | 9       | 17     | N/A      | 9      | 13     | N/A     |
| English  | 6      | 100     | 28     | 7        | 26     | 19     | 14      |
| German   | 6      | 15      | 100    | N/A      | 5      | 8      | 4       |
| Japanese | N/A    | 2       | N/A    | 100      | 8      | N/A    | N/A     |
| Korean   | 1      | 5       | 2      | 4        | 100    | 1      | 3       |
| Polish   | 6      | 12      | 9      | N/A      | 5      | 100    | 15      |
| Russian  | N/A    | 11      | 7      | N/A      | 11     | 19     | 100     |

**eLinguistics** [23] calculates a genetic proximity score between two languages by comparing the consonants contained in a predefined set of words, while taking the order in which these consonants appear in the words into account. This gives information about the direct relatedness of the compared languages. The quantification of the consonant relationships is based on the work by Brown et al. [71].

However, the method seemed to become more prone to errors as the distance between the two compared languages increased as more and more accidental similarities in consonants are introduced. While being simple in formulation and completely disregarding semantic, morphological, and syntactic similarity, the similarity values seemed to be in line with our intuition and the other metrics used in this study. The metric is also easily accessible through a web service [7]. The values for our proposed languages can be found in Table 2.

**EzGlot** [24] uses lexical similarity (similarity of vocabularies) of a pair of languages. The Ezglot linguistic similarity measure is calculated using lexical similarity between the two compared languages, while also considering the number of words these languages share with other languages. This way, it is possible to calculate the similarity between the two languages in relation to similarities to all other languages. Also, as the number of words the languages share with other languages are also taken into account, because of this calculation, metric becomes asymmetric between the language pairs, which matches well with the fact that the mutual intelligibility of languages is also considered asymmetric [72, 21].

The calculation formula and a pre-calculated similarity matrix for EzGlot are also easily accessible on the project's website [8]. However, the similarity matrix is missing values for multiple languages, for example Japanese which is used in our study, hindering its usability. Also, the authors do not reveal the source of their data, leaving the quality of the results under question and making it more difficult to fill the gaps in the similarity matrix. A sample of this similarity matrix containing our proposed languages is shown in Table 3.

Additionally we were planning to use the similarity metric STL [22], which utilizes multiple aspects of languages. This would have combined Semantic, Terminological (lexical) and Linguistic (syntactic) similarity of languages into a single metric and according to the authors, outperformed previous similarity metrics that were using only a single one of these aspects [45, 46]. However, the ambiguous presentation of the calculation formula made it impossible to be utilized in this research. Instead, we propose a novel similarity metric quantified from the World Atlas of Language Structures (WALS).

---

[7] http://www.elinguistics.net/Compare_Languages.aspx
[8] https://www.ezglot.com/most-similar-languages.php

**Table 4**
WALS distance metric

|  | Danish | English | German | Japanese | Korean | Polish | Russian |
|---|---|---|---|---|---|---|---|
| **Danish** | 0.000 | 0.098 | 0.086 | 0.252 | 0.212 | 0.154 | 0.152 |
| **English** | 0.098 | 0.000 | 0.149 | 0.306 | 0.240 | 0.155 | 0.152 |
| **German** | 0.086 | 0.149 | 0.000 | 0.332 | 0.292 | 0.181 | 0.179 |
| **Japanese** | 0.252 | 0.306 | 0.332 | 0.000 | 0.118 | 0.295 | 0.277 |
| **Korean** | 0.212 | 0.240 | 0.292 | 0.118 | 0.000 | 0.232 | 0.229 |
| **Polish** | 0.154 | 0.155 | 0.181 | 0.295 | 0.232 | 0.000 | 0.077 |
| **Russian** | 0.152 | 0.152 | 0.179 | 0.277 | 0.229 | 0.077 | 0.000 |

### 4.3. The World Atlas of Language Structures

The World Atlas of Language Structures (WALS) project [25] consists of a database that catalogs phonological, word semantic and grammatical knowledge for over 2,662 languages in over 200 language families. The database consists of 192 different features as of the time of writing (March 2022). However, not all of the features are documented for all of the available languages, for example English has around 150 documented features. This number goes down for less studied languages, for example, Danish has only 58 documented features available[9]. All languages and features considered, this sums up to a total of over 58,000 data points, which means the database is only 12% populated. Even major languages are lacking values for multiple features. For example, English is missing around 25% of the total features. This data sparsity is our main concern in quantifying the database into a language similarity metric as using more languages means having less common features among them.

Another one of the goals of this study was to create a linguistic similarity metric, that would take multiple aspects of a language into account instead of only using a single feature, from the WALS database. To accomplish this we downloaded a snapshot and scanned through the database. Then we selected all of the features that would have a defined value in all of the proposed languages (English, German, Danish, Polish, Russian, Japanese and Korean). This resulted in a total of 42 common features among the languages. Next we looked at the possible values each feature can take and converted the values for each feature to a numeric scale from zero to one in the relative order. To do this, we assumed that the order in which the possible values are presented in the WALS database roughly represents the order of their similarity. After converting the values, we used them to compare each language pair and calculate an average of the euclidean distances between all of the features for all of the possible combinations of languages, resulting in a symmetric distance metric. The finished distance matrix is shown in Table 4.

## 5. Experiments

### 5.1. Setup

We fine-tuned both of the models (mBERT, XLM-R) with all of the proposed languages (English, German, Danish, Polish, Russian, Japanese and Korean), producing a total of 14 models. The fine-tuned models were then evaluated with test datasets from each of the proposed languages in order to calculate the cross-lingual transfer performances. The models were evaluated using a macro F1-score. After evaluating the models, we studied the correlation between classifier performance and language similarity using the previously introduced linguistic similarity metrics. Specifically, we calculated both Pearson's and Spearman's correlation coefficients between the models and the linguistic similarity metrics. The training of the classifiers was done using PyTorch and the Transformers library on an Nvidia GTX 1080Ti.

---

[9]Some languages have even less features described, e.g., spoken in Guatemala Chuj language has only 29, while spoken in Indonesia Kutai has only one feature described.

**Table 5**
Classification scores (F-score) for Multilingual BERT

|  |  | TARGET | | | | | | |
|---|---|---|---|---|---|---|---|---|
|  |  | Danish | English | German | Japanese | Korean | Polish | Russian |
|  | Danish | 0.75 | 0.54 | 0.50 | 0.37 | 0.40 | 0.51 | 0.55 |
|  | English | 0.53 | 0.77 | 0.41 | 0.33 | 0.34 | 0.44 | 0.41 |
| SOURCE | German | 0.57 | 0.56 | 0.70 | 0.48 | 0.45 | 0.63 | 0.72 |
|  | Japanese | 0.50 | 0.53 | 0.46 | 0.88 | 0.49 | 0.49 | 0.51 |
|  | Korean | 0.43 | 0.44 | 0.33 | 0.45 | 0.95 | 0.38 | 0.50 |
|  | Polish | 0.51 | 0.48 | 0.41 | 0.33 | 0.34 | 0.83 | 0.58 |
|  | Russian | 0.50 | 0.47 | 0.61 | 0.64 | 0.61 | 0.60 | 0.90 |

**Table 6**
Classification scores (F-score) for XLM-RoBERTa

|  |  | TARGET | | | | | | |
|---|---|---|---|---|---|---|---|---|
|  |  | Danish | English | German | Japanese | Korean | Polish | Russian |
|  | Danish | 0.75 | 0.67 | 0.49 | 0.39 | 0.39 | 0.56 | 0.57 |
|  | English | 0.58 | 0.81 | 0.43 | 0.46 | 0.41 | 0.47 | 0.45 |
| SOURCE | German | 0.66 | 0.66 | 0.73 | 0.55 | 0.52 | 0.64 | 0.71 |
|  | Japanese | 0.56 | 0.57 | 0.47 | 0.90 | 0.61 | 0.55 | 0.65 |
|  | Korean | 0.43 | 0.49 | 0.34 | 0.52 | 0.95 | 0.39 | 0.48 |
|  | Polish | 0.63 | 0.60 | 0.50 | 0.46 | 0.50 | 0.84 | 0.78 |
|  | Russian | 0.57 | 0.53 | 0.66 | 0.72 | 0.76 | 0.65 | 0.91 |

### 5.2. Classification Results

We fine-tuned the multilingual transformer models with the offensive language datasets described earlier. Each model was fine-tuned only on a single language before evaluation. The classifier evaluation results were shown in Tables 5 and 6.

From the results, it is clear that XLM-R outperformed mBERT, as its scores are higher across the board. Also, the scores are obviously highest when using the same language as source and target. The second highest scores are usually by the languages in the same language families (English, German, Danish - Germanic; Polish, Russian - Slavic; Japanese, Korean - Koreano-Japonic).

As can be seen from Tables 5 and 6, English was generally the worst language to use as the source, having low scores with all languages but itself. German and Danish worked best as the source language for English. In general, German worked well as a source language but was hard to generalize on by other languages. For example, both English and Danish worked better as a source for Polish and Russian than German. In addition, German worked especially well as a source language for the Slavic languages (Polish and Russian). Also, Russian had the highest zero-shot performance as the source language for German. Also, German was the best source for Danish. Polish received a good score as the source language for Russian. However, Russian did not do so well as the source for Polish, being equalled by German. Interestingly, Russian had good scores as the source language for Japanese and Korean, a feature that no other language had, outperforming both Japanese and Korean when used as source languages for one another. Specifically, Russian to Japanese yielded a score of 0.64 while Korean to Japanese was only 0.45, and the score for Russian to Korean was 0.61 while Japanese to Korean was 0.49 with XLM-R. Generally, Japanese and Korean were the hardest target languages and interestingly, did not score well as a language pair despite being classified in the same language family.

### 5.3. Correlation with Linguistic Similarity

We calculated Pearson's and Spearman's correlation coefficients ($p$-value) between the classification results of the two classifiers and each of the three proposed linguistic similarity metrics (EzGlot, eLinguistics, WALS). As the similarity matrix for EzGlot was not fully populated, we needed to ignore the scores for the missing language pairs during the calculation of the correlations for this particular similarity metric. The results can be seen in Tables 7 and 8 for Pearson's and Spearman's correlation coefficients respectively.

**Table 7**
Pearson's correlation coefficient for classifier scores and linguistic similarity metrics

|  | XLM-R | | mBERT | |
| --- | --- | --- | --- | --- |
|  | $\rho$ | p-value | $\rho$ | p-value |
| **WALS** | -0.674 | 0.001 | -0.713 | 0.001 |
| **EzGlot** | 0.720 | 0.001 | 0.801 | 0.001 |
| **eLinguistics** | -0.713 | 0.001 | -0.736 | 0.001 |

**Table 8**
Spearman's correlation coefficient for classifier scores and linguistic similarity metrics

|  | XLM-R | | mBERT | |
| --- | --- | --- | --- | --- |
|  | $\rho$ | p-value | $\rho$ | p-value |
| **WALS** | -0.599 | 0.001 | -0.615 | 0.001 |
| **EzGlot** | 0.506 | 0.001 | 0.494 | 0.001 |
| **eLinguistics** | -0.654 | 0.001 | -0.666 | 0.001 |

**Table 9**
Pearson's correlation coefficient after removing the same source-target language pairs

|  | XLM-R | | mBERT | |
| --- | --- | --- | --- | --- |
|  | $\rho$ | p-value | $\rho$ | p-value |
| **WALS** | -0.359 | 0.020 | -0.368 | 0.016 |
| **EzGlot** | 0.011 | 0.953 | -0.040 | 0.829 |
| **eLinguistics** | -0.438 | 0.004 | -0.421 | 0.006 |

As can be seen from the results, Pearson's correlation is strong with all of the proposed similarity metrics. Also, the p-value is less than 0.05 in all of the cases, showing that the results are statistically significant. EzGlot's similarity metric has the strongest correlation with $\rho = 0.720$ for XLM-R and $\rho = 0.801$ for mBERT. This is followed by eLinguistics and WALS with the absolute correlation in the range of 0.67 to 0.73. Spearman's correlation is slightly lower, being in the moderate-strong range with all of the metrics, with the p-value also being less than 0.05. The strongest correlation is by eLinguistics with an absolute correlation of $\rho = 0.654$ for XLM-R and $\rho = 0.666$ for mBERT. This is followed by eLinguistics and WALS with the absolute correlation in the range of 0.49 to 0.62. Also, the correlations were generally slightly stronger with mBERT than with XLM-R.

However, after removing the same source-target language pairs and leaving only the zero-shot classification results, the correlations changed drastically. This is shown on Tables 9 and 10 for Pearson's and Spearman's correlation coefficients, respectively. There were two changes. First, EzGlot's similarity metric plummeted down from having the strongest correlation with Pearson's to showing no correlation at all, both correlation coefficients showing a value near zero, and losing statistical significance. Second, the correlations for eLinguistics and WALS also fell from strong to moderate, standing now in the range of 0.35 to 0.44 for Pearson's and 0.37 to 0.48 for Spearman's correlation coefficient. The p-values also increased slightly, but remained under 0.05, keeping statistical significance. Also, the correlation of eLinguistics stayed stronger than that of WALS despite the other changes.

## 6. Discussion
### 6.1. Transfer Language Performance

The fact that XLM-R outperformed mBERT matches our expectations, as it also did so on a variety of benchmark tasks [73, 74]. The reasons are most likely that XLM-R is a true cross-lingual model and has a vastly larger vocabulary size than mBERT. The highest cross-lingual transfer scores (Tables 5 and 6) were usually by languages from the same language families as the source language. This matches with the typical intuitive selection process when selecting the transfer source language. However, this was not always the case and one relying purely on selecting the languages from the same language family will lead to diminished transfer performance in some cases. A good example of this is when the target language is German. One could expect the best transfer languages being Danish and English, but actually both Polish and Russian had a higher transfer performance despite of being from the Slavic language family, not Germanic. This could be due to the differences in grammatical complexity. Both Danish and English have relatively simple grammar compared to German, which could leave them unable to generalize on the more sophisticated German language. On the other hand, the grammar of Polish and Russian is even more complex, which could negate this issue, allowing them to generalize better. Also, German, Polish and Russian all have synthetic morphology, which could play a role in their ability to generalize well on each other. Furthermore, the historical mutual influence between Germans, Poles and Russians could be a factor here. Looking at the scores, it can be noted that German is a good source for both Germanic and Slavic languages.

**Table 10**
Spearman's correlation coefficient after removing the same source-target language pairs

|  | XLM-R | | mBERT | |
| --- | --- | --- | --- | --- |
|  | $\rho$ | p-value | $\rho$ | p-value |
| **WALS** | -0.377 | 0.014 | -0.395 | 0.010 |
| **EzGlot** | 0.129 | 0.480 | 0.100 | 0.584 |
| **eLinguistics** | -0.465 | 0.002 | -0.475 | 0.001 |

English on the other hand was one of the worst, if not the worst language to use as the transfer source overall. It had a poor performance even in its own language family, probably due to its simplicity when compared to both Danish and German. Also, English is heavily influenced by French, further distancing it from the other Germanic languages. Furthermore, the differences in morphology could be a factor. The analytic nature of English could be a reason why it cannot generalize on fusional languages like German. Danish, which is also an analytic language had a better generalization for German probably due to its otherwise closer ties to the German language (eg., mutual influence). These results show that other languages should be considered over English as the cross-lingual transfer source if available.

Interestingly, Russian achieved a high score as the transfer language source for both Korean and Japanese. Which is different from any of the other languages included in this study. The reason could be in the shared morphological features, specifically the fact that all of the three languages are agglutinative. Furthermore, all of the three languages contain distinct registers, specifically for expressing different politeness levels. However, as this could be heavily related to the topic of offensive language identification or to the properties of these specific datasets and might not be applicable to other fields, this needs to be confirmed in the future on other datasets and, preferably, different tasks as well.

Also, Korean and Japanese did not work as well with each other, contrary to how we were expecting, and were clearly outperformed by Russian, even though they are more similar with one another than with any of the other languages used in this study. This as well could be related to the properties of the datasets or to the topic of offensive language identification itself and will have to be verified in the future.

Furthermore, when looking at the source languages individually, we could see a trend of the transfer performance being better when transferring to more similar languages in all cases except Russian, as it tended to have an exceptionally high transfer performance to both Japanese and Korean when compared to other languages. For all of the other languages, the performance clearly decreases as the distance between the languages increases. This can be seen from Figure 1.

The sizes of other datasets vary from around 3,000 samples to almost 200,000 samples. Also, the percentage of harmful samples in each dataset varies greatly. The English dataset has only 7% of harmful samples while Japanese and Korean have around 50%. In order to determine, whether the dataset size and balance had an effect on the results or not, we decided to further analyze the effectiveness of each dataset as the transfer source. In order to measure the effect, we calculated the correlation between the average transfer scores of each dataset and the dataset size/balance by taking a base-10 logarithm of the dataset size, multiplied by the proportion of harmful samples in the dataset. We had to take the logarithm of the sample size, because otherwise its weight would become too large compared to the harmful ratio, which is bound between 0 and 1. The results are shown in Table 11.

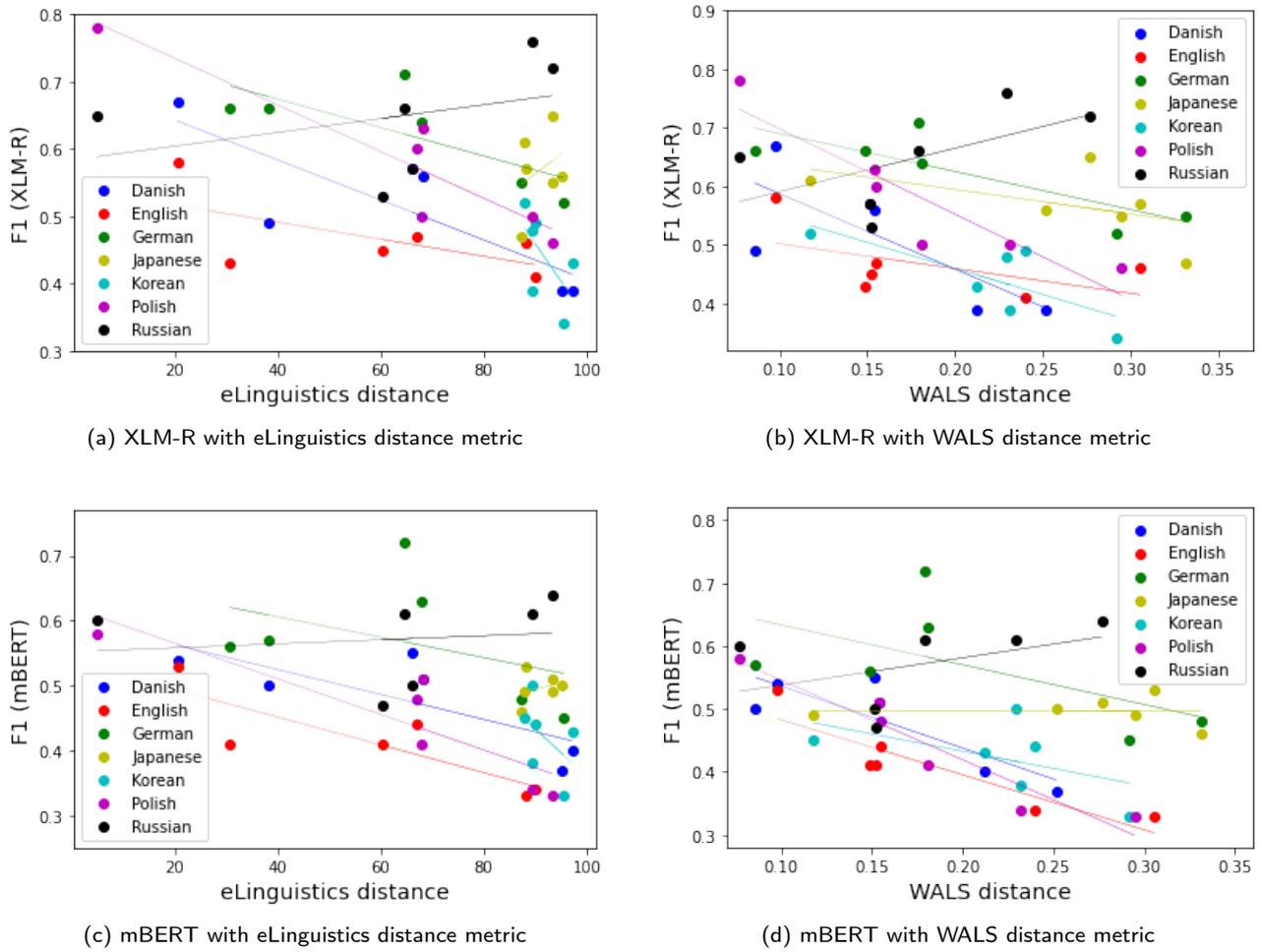

**Figure 1:** Performance trends for source languages for cross-lingual transfer

As can be seen from the results, mBERT and XLM-R show no correlation between the average performance of the classifier as the source language and the size/balance of the dataset. With this we can more safely say that the results should not be biased by the differences in the size and balance of the datasets.

### 6.2. Analysis of Specific Examples

In order to clarify the characteristics of the classification models, we chose a number of example sentences from the English cyberbullying test dataset based on the prediction results and confidence (strength of prediction by the model). We inspected the results and tried to reason why the model made the decisions, concentrating on potential points of failure. The texts were chosen by taking three properties into account, confidence (low/high), label (0/1) and prediction result (incorrect/correct), resulting in eight examples. The results can be found in Table 12.

Looking at the results, the low confidence texts seem to have two factors in common. First, they contain many slang words, spoken contractions and typos. These words are most likely not contained in the pretrained model's vocabulary and thus make the texts harder to classify. Second, the annotations of these texts seem more ambiguous and could be interpreted in different ways. For example, looking at the third example it is not impossible to see why it has been labeled as it is (revealing someone's location by city and using vulgar language), but one could also argue that these alone are not enough to label the text as cyberbullying. Also, the expression used in the fifth text could mean either sexual demand or be a sarcastic comment about a frustrating situation. Although here the context implies it is sexual.

The mistakes the model made with high confidence are more difficult to reason. For example, the second example does not imply anything cyberbullying related but the model still predicted it as so. In the future, we plan to investigate the attentions put on each tokens to be able to produce a working theory of why the model makes such decisions. Similar thing applies to the fourth sentence, although the label and the prediction are the opposite. The sixth and eight examples were correctly classified with high confidence and they clearly represent their assigned labels.

**Table 11**
Upper: average F1 and dataset size and balance statistics, lower: Pearson's correlation coefficient for average F1 and size/balance

|  | mBERT | XLM-R | Total samples (nS) | Harmful ratio (Hr) | log(nS)*Hr |
|---|---|---|---|---|---|
| **Danish** | 0.48 | 0.51 | 3,289 | 0.13 | 0.45 |
| **English** | 0.41 | 0.47 | 12,772 | 0.07 | 0.29 |
| **German** | 0.57 | 0.62 | 8,407 | 0.34 | 1.32 |
| **Japanese** | 0.5 | 0.57 | 6,434 | 0.50 | 1.91 |
| **Korean** | 0.42 | 0.44 | 189,995 | 0.47 | 2.50 |
| **Polish** | 0.44 | 0.58 | 34,953 | 0.21 | 0.96 |
| **Russian** | 0.57 | 0.65 | 14,412 | 0.33 | 1.39 |
| **Pearson $\rho$** | 0.115 | 0.111 | | | |
| **p-value** | 0.977 | 0.812 | | | |

**Table 12**
Example sentences, predictions and confidence values (XLM-R)

| Text | Label | Pred (en) | Conf (en) | Pred (da) | Conf (da) | Pred (de) | Conf (de) | Pred (pl) | Conf (pl) | Pred (ru) | Conf (ru) | Pred (ja) | Conf (ja) | Pred (ko) | Conf (ko) |
|---|---|---|---|---|---|---|---|---|---|---|---|---|---|---|---|
| I GUNA PEEEEEEEE MY PANTS CUZ OV U go wee then sori | 0 | 1 | 0.52 | 0 | 0.71 | 0 | 0.58 | 0 | 0.91 | 1 | 0.75 | 0 | 0.84 | 0 | 0.64 |
| Are you a nun Ummm Nopee... Dont Thinkk Soo.. :) | 0 | 1 | 0.83 | 0 | 0.75 | 0 | 0.83 | 0 | 0.94 | 1 | 0.62 | 0 | 0.83 | 0 | 0.76 |
| brittany is from MAYODAN abbie aint from ellisboro either.dumbass....and the lst 2 aintchur girls they dont evn lik u. I was naming people who live out in that area ha yeahh okayy. I really don't care what some bitch thinks about me (: | 1 | 0 | 0.55 | 1 | 0.70 | 1 | 0.69 | 0 | 0.92 | 1 | 0.77 | 1 | 0.67 | 0 | 0.51 |
| what do you think about men who like to be dominated and ridiculed in bed? LMAO suckers ! ! | 1 | 0 | 0.93 | 1 | 0.57 | 1 | 0.65 | 1 | 0.69 | 1 | 0.79 | 1 | 0.76 | 0 | 0.89 |
| —fuck me— hah... nah stranger;) | 1 | 1 | 0.64 | 1 | 0.63 | 0 | 0.81 | 0 | 0.95 | 1 | 0.57 | 0 | 0.77 | 0 | 0.83 |
| you can suck the dick bitch dont try me little girl dany shut the fuck upppp BITCH...r r | 1 | 1 | 0.87 | 1 | 0.77 | 1 | 0.62 | 0 | 0.82 | 1 | 0.79 | 1 | 0.85 | 1 | 0.81 |
| i love how the people who talk shit are all anonymous. they're scared of youu (:r lolll. lofl i fukin noee rytee;] | 0 | 0 | 0.50 | 1 | 0.69 | 0 | 0.53 | 0 | 0.91 | 1 | 0.74 | 0 | 0.57 | 1 | 0.53 |
| Shouldn't the opposite of shut up be shut down? i guesss so. ha | 0 | 0 | 0.94 | 0 | 0.88 | 0 | 0.83 | 0 | 0.95 | 0 | 0.76 | 0 | 0.81 | 0 | 0.78 |

Looking at the predictions and confidences of the models fine-tuned in other languages than English, the situation is quite different. It looks like the other models performed better on these texts than the English model. For example, the Danish, German and Japanese models classified seven out of these eight texts correctly. Also, the confidences seem to vary a lot, with both close and distant languages. In the future, we plan to take a deeper look into the cross-lingual classification results and investigate, what could be the possible causes for the model's behaviour.

**Table 13**
Target languages, best sources and their similarity ranks for eLinguistics and WALS metrics

| Target | Best source | eLinguistics | WALS |
| --- | --- | --- | --- |
| Danish | German | 2 | 1 |
| English | Danish | 1 | 1 |
| German | Russian | 3 | 3 |
| Japanese | Russian | 4 | 4 |
| Korean | Russian | 2 | 3 |
| Polish | Russian | 1 | 1 |
| Russian | Polish | 1 | 1 |

### 6.3. Analysis of Linguistic Similarity Metrics

The correlation of EzGlot's similarity metric was higher than those of eLinguistics' or WALS' as can be noted from Tables 7 and 8. Being based on lexical similarity, it would suggest that the used multilingual models heavily rely on lexical information. However, when considering only the zero-shot classification results (Tables 9 and 10), EzGlot's similarity metric changed to showing no correlation with the transfer performance at all. This shows that they do not rely only on lexical features and that other linguistic features need to be considered when choosing the source language for cross-lingual transfer.

To our surprise, the correlation of eLinguistics' metric was higher than the correlation of the multidomain metric we quantified from WALS despite of being calculated only by comparing a predetermined set of phonetic consonants. Possibly, our choice of including only the features common among all of the proposed languages could have caused too many irrelevant features to be included. This might have resulted in bias in the metric calculation. In the future, we will aim for a better quantification of the WALS database in order to develop an even more effective and comprehensive similarity metric, also by incorporating the other two metrics.

However, even though having the highest correlation, the eLinguistics metric has its weaknesses due to being based on only one aspect of language. Looking at Table 2 one can see that eLinguistics shows Japanese being very distant from Korean, being at the same level as Polish and Russian, which is in fact not true due to similarities in vocabulary and grammar between Japanese and Korean. The WALS metric on the other hand is obviously more robust to errors like this as can be seen from Table 4. This is most likely thanks to it being based on multiple linguistic features instead of only one as is the case with eLinguistics metric. Table 13 shows that on average the metrics look equally good if they were used for transfer language selection. Only difference being that when using WALS, the best transfer option was chosen more often.

The fact that the similarity metrics of eLinguistics and WALS correlated with transfer language efficacy means that they can be used for the selection process. So, instead of making a decision based on intuition or simply choosing any language from the same language family, one can check the similarity of the target language with high-resource languages that have proper data available and make a more informed and effective decision, at least for offensive language identification. This allows for more efficient model development.

### 6.4. Ethical Considerations

Being able to choose an optimal transfer language can greatly aid in the task of detecting harmful language like hate speech and cyberbullying, especially when dealing with low-resource languages. The fact that there are thousands of languages used every day in online communication and social media, of which only a small fraction have proper data to train the detection models on, shows the amount of potential this method has. This will make it possible to detect offensive content as early and effectively as possible to prevent its serious consequences and control its spread. The method ultimately aids in reducing the damages abusive and harmful content causes to the society. Also, it can reduce the human effort required to keep offensive content like cyberbullying and hate speech at bay, release the society's resources for development in other fields and ease the burden on those who have to deal with this serious problem in any way.

**Table 14**
Upper: Pretraining corpus size and average classifier performance (F1), lower: Pearson's and Spearman's correlation coefficient for pretraining corpus size and average F1

| Language | Size (GB) | F1 (XLM-R) |
|---|---|---|
| **English** | 300.8 | 0.47 |
| **Russian** | 278.0 | 0.65 |
| **Japanese** | 69.3 | 0.57 |
| **German** | 66.6 | 0.62 |
| **Korean** | 54.2 | 0.44 |
| **Danish** | 45.8 | 0.51 |
| **Polish** | 44.6 | 0.58 |
| **Pearson** | $\rho$: 0.085 | p-value: 0.86 |
| **Spearman** | $\rho$: 0.071 | p-value: 0.88 |

To take a look at the ethicality of the used transformer models, we need to inspect the data used for their pretraining. The corpora used for pretraining are from Wikipedia (mBERT) and CommonCrawl (XLM-R) [27]. These differ greatly in domain as Wikipedia consists only of well structured documents written in formal language, whereas CommonCrawl, being basically a snapshot of the Web, might contain almost anything from structured text (Wikipedia) to more natural texts like blogs. This means that while Wikipedia is already mostly internally detoxified due to its ethical guidelines and moderation, the CommonCrawl data most likely contains also unethical matter not only because of the inclusion of more natural texts like blogs or product reviews, but also as the result of media bias induced by the addition of news data. Paradoxically, the fact that XLM-R is also pretrained on possibly toxic content could be a contributing factor to its higher performance in the cyberbullying detection task. In order to investigate the effect of including possibly unethical material in model pretraining, we calculated the correlation between the proportion of possibly unethical text (non-Wikipedia) in the corpus and our classifier performance.

As the exact amounts of Wikipedia data used in the pretraining were not available, we ranked the proposed languages based on the approximate proportions [27] of unethical text and calculated Spearman's rank correlation coefficient between the approximate proportions and the classifier scores[10]. This resulted in Spearman $\rho = 0.036$, meaning that we could not find a correlation between using possibly unethical data in pretraining and the performance of a fine-tuned model. However, due to the limited scope of this research, further study is required to investigate the effect of including unethical text in the pretraining process.

In addition to containing texts from different domains, the corpora also holds different amount of data for different languages. This could cause initial bias in the language coverage, which could also have an impact on the performance. Looking at the amount of data used from our proposed languages, the pretraining corpora sizes with XLM-R vary from 300GB (English) to 45GB (Polish). In order to determine whether the results are biased by the amount of pretraining data, we calculated the correlation between the pretraining corpus size and classifier performance. The results were shown in Table 14.

In the case of our proposed languages, we were not able to notice that the pretraining corpus size influenced the results. However, further research is required to investigate the possible bias in pretrained multilingual models considering the used pretraining corpora sizes for each language.

### 6.5. Limitations

Naturally, the pretraining data size should have an effect to the classifier's performance even though we didn't find any correlation. One of the factors in the overall high performance of the Russian model could be the high pretraining data size. However, the English model performed poorly despite of being pretrained on the largest corpus. Unfortunately it is impossible to account for this unless we pretrain the models ourselves. The point of these experiments was to concentrate on optimizing the usage of existing resources, but we consider pretraining the models with similar corpora for the next steps of the research in order to have a more uniform setup and to reduce the amount of confounding factors.

---

[10]Here, we used only Spearman's rank correlation coefficient without Pearson's correlation coefficient, since, although we were able to deduce the relative approximate proportions, the exact numbers were not mentioned in the literature.

Another factor is the differences between the fine-tuning datasets. The dataset size and the ratio of positive and negative class being one that could bias the results. However, we couldn't find a correlation between these factors and the classifier performance. Another difference comes with the dataset domains. Cyberbullying, hate speech and toxic language have their innate differences and could impact the transfer performance in a cross-domain case [75]. Unfortunately fixing these issues would require us to collect and annotate all of the datasets ourselves as quality datasets are already scarce when considering many of the used languages. Also, the main goal of the paper was to aid in the creation of offensive language detection models for especially low-resource languages using what is currently available. In the future, when one becomes available, we are planning to repeat the experiments on a fully cross-lingual offensive language dataset created as uniformally as possible apart from language.

### 6.6. Future Research

In the future, we are planning to re-quantify the WALS database in order to develop an even more effective and comprehensive similarity metric. In the current implementation, many features were cut out due to the data being too sparse. One option here would be to calculate the features separately for each language pair. This would allow us to capture more features per language, but would lead to inconsistencies when comparing languages as a different amount of features would be used. Another solution would be to determine which features have the strongest correlation with the cross-lingual transfer performance and then create the metric based on those features. Also, it would be a great aid if the WALS project received more attention and the feature matrix became more populated.

We are hypothesizing that this method could be useful as a general method also for other Natural Language Processing tasks outside of offensive language identification. This would help the model developer to make a more effective and justifiable decision instead of relying on intuition or simply choosing a language from the same family. We need to confirm the effectiveness of the selection method also for other tasks like sentiment analysis [76, 77], dependency parsing [78, 67, 79], named entity recognition [67, 80] and machine translation [81, 82]. This will be done using more general benchmark datasets [73, 74].

We plan to implement the method in the development of a multilingual cyberbullying detection application. With a proper transfer language selection procedure, we are able to deal with some of the difficulties encountered earlier with low-resource languages. Furthermore, the linguistic features described in WALS could help us find new insights about the features of offensive content in order to further practical research on cyberbullying and hate speech and ways of their mitigation.

## 7. Conclusions

In this study we studied the selection of transfer languages for automatic hate speech detection. We demonstrated the effectiveness of cross-lingual transfer learning for zero-shot offensive language identification on a target language. This way it is possible to leverage existing data from higher-resource languages in order to improve the performance of languages lacking proper data. We showed that there is a strong correlation between our proposed linguistic similarity metrics and the cross-lingual transfer performance. As the languages get more distant, the transfer performance decreases. This makes it possible to choose an optimal transfer language by comparing the similarity of languages instead of relying on intuition. As shown by our experiments, choosing languages from the same language family is not always the best option. Instead, one should use different linguistic features to compare the languages before selection and base the choice on a linguistic similarity metric instead. Our experiments also showed that lexical information alone is not enough to determine the optimal transfer languages.

We also showed that it is possible to achieve good performance on the target language in a zero-shot cross-lingual transfer setting. This helps in developing better detection systems for offensive language identification, especially when dealing with low-resource languages. This is particularly important because of the severity of the problem and the fact that social media is used in thousands of languages, of which only a small fraction even have proper data to train the detection models on.

Lastly, we developed a novel linguistic similarity metric consisting of various linguistic features by using the WALS database. Our proposed method did not show the strongest correlation with the transfer performance, but it still showed potential as a metric that could be useful for the selection process, especially if given a more refined or inclusive feature set. In the future, we will aim for a better quantification of the WALS database in order to develop an even more effective and comprehensive linguistic similarity metric.

The proposed method for cross-lingual transfer language selection could also be useful as a general method for other Natural Language Processing tasks, not only for harmful online content detection. In the near future, we plan to confirm the effectiveness of the selection method also for other NLP tasks like sentiment analysis and machine translation. This will be done using more general benchmark datasets.

## CRediT authorship contribution statement

**Juuso Eronen:** Conceptualization, Methodology, Software, Writing - Original Draft, Writing - Review & Editing. **Michal Ptaszynski:** Conceptualization, Methodology, Supervision, Data Curation. **Fumito Masui:** Conceptualization, Methodology, Supervision, Data Curation. **Masaki Arata:** Data Curation. **Gniewosz Leliwa:** Data Curation. **Michal Wroczynski:** Data Curation.

## References

[1] Sameer Hinduja and Justin Patchin. Bullying, cyberbullying, and suicide. *Archives of suicide research : official journal of the International Academy for Suicide Research*, 14:206–21, 07 2010.

[2] G. Sarna and M.P.S. Bhatia. Content based approach to find the credibility of user in social networks: an application of cyberbullying. *Int. J. Mach. Learn. and Cyber*, 8:677–689, 2015.

[3] Michal Ptaszynski and Fumito Masui. *Automatic Cyberbullying Detection: Emerging Research and Opportunities*. IGI Global, 2018.

[4] Bertie Vidgen and Leon Derczynski. Directions in abusive language training data, a systematic review: Garbage in, garbage out. *PLOS ONE*, 15(12):1–32, 12 2021.

[5] April Kontostathis, Kelly Reynolds, Andy Garron, and Lynne Edwards. Detecting cyberbullying: Query terms and techniques. In *Proceedings of the 5th Annual ACM Web Science Conference*, WebSci '13, page 195–204, New York, NY, USA, 2013. Association for Computing Machinery.

[6] Karthik Dinakar, Birago Jones, Catherine Havasi, Henry Lieberman, and Rosalind Picard. Common sense reasoning for detection, prevention, and mitigation of cyberbullying. *ACM Transactions on Interactive Intelligent Systems*, 2, 09 2012.

[7] Michal Ptaszynski, Pawel Dybala, Tatsuaki Matsuba, Fumito Masui, Rafal Rzepka, and Kenji Araki. Machine learning and affect analysis against cyber-bullying. In *Linguistic And Cognitive Approaches To Dialog Agents Symposium*, 03 2010.

[8] Michal Ptaszynski, Pawel Dybala, Tatsuaki Matsuba, Fumito Masui, Rafal Rzepka, Kenji Araki, and Yoshio Momouchi. In the service of online order: Tackling cyber-bullying with machine learning and affect analysis. *International Journal of Computational Linguistics Research*, 1(3):135–154, 2010.

[9] Michal Ptaszynski, Fumito Masui, Yasutomo Kimura, Rafal Rzepka, and Kenji Araki. Extracting patterns of harmful expressions for cyberbullying detection. In *7th Language and Technology Conference (LTC'15), The First Workshop on Processing Emotions*, 11 2015b.

[10] Sameer Hinduja and Justin W Patchin. Cyberbullying research center. https://cyberbullying.org/, 2021.

[11] Glen Bull. The always-connected generation. *Learning and Leading with Technology*, 38:28–29, 2010.

[12] María Antonia Paz, Julio Montero-Díaz, and Alicia Moreno-Delgado. Hate speech: A systematized review. *SAGE Open*, 10(4):2158244020973022, 2020.

[13] Justin W. Patchin and Sameer Hinduja. Bullies move beyond the schoolyard: A preliminary look at cyberbullying. *Youth Violence and Juvenile Justice*, 4(2):148–169, 2006.

[14] MEXT. 'netto-jō no ijime' ni kansuru taiō manyuaru jirei shū (gakkō, kyōin muke) ["bullying on the net" manual for handling and collection of cases (for schools and teachers)] (in japanese). *Ministry of Education, Culture, Sports, Science and Technology (MEXT)*, 2008.

[15] Michal Ptaszynski, Fumito Masui, Taisei Nitta, Suzuha Hatakeyama, Yasutomo Kimura, Rafal Rzepka, and Kenji Araki. Sustainable cyberbullying detection with category-maximized relevance of harmful phrases and double-filtered automatic optimization. *International Journal of Child-Computer Interaction*, 8, 08 2016.

[16] Tharindu Ranasinghe and Marcos Zampieri. Multilingual offensive language identification with cross-lingual embeddings. In *Proceedings of the 2020 Conference on Empirical Methods in Natural Language Processing*, page 5838–5844, November 2020.

[17] Tharindu Ranasinghe and Marcos Zampieri. Multilingual offensive language identification for low-resource languages, 2021.

[18] Irina Bigoulaeva, Viktor Hangya, and Alexander Fraser. Cross-lingual transfer learning for hate speech detection. In *Proceedings of the First Workshop on Language Technology for Equality, Diversity and Inclusion*, pages 15–25, Kyiv, April 2021. Association for Computational Linguistics.

[19] Saurabh Gaikwad, Tharindu Ranasinghe, Marcos Zampieri, and Christopher M. Homan. Cross-lingual offensive language identification for low resource languages: The case of marathi, 2021.

[20] Ryan Cotterell and Georg Heigold. Cross-lingual character-level neural morphological tagging. In *Proceedings of the 2017 Conference on Empirical Methods in Natural Language Processing*, pages 748–759, Copenhagen, Denmark, September 2017. Association for Computational Linguistics.

[21] Charlotte Gooskens, Vincent J. van Heuven, Jelena Golubović, Anja Schüppert, Femke Swarte, and Stefanie Voigt. Mutual intelligibility between closely related languages in europe. *International Journal of Multilingualism*, 15(2):169–193, 2018.

[22] Nitish Aggarwal, Tobias Wunner, Mihael Arčan, Paul Buitelaar, and Seán O'Riain. A similarity measure based on semantic, terminological and linguistic information. In *Proceedings of the 6th International Workshop on Ontology Matching*, 01 2011.

[23] Vincent Beaufils and Johannes Tomin. Stochastic approach to worldwide language classification: the signals and the noise towards long-range exploration, Oct 2020.

[24] Lazar Kovacevic, Vladimir Bradic, Gerard de Melo, Sinisa Zdravkovic, and Olga Ryzhova. Ezglot. https://www.ezglot.com/, 2021.

[25] Matthew S. Dryer and Martin Haspelmath, editors. *WALS Online*. Max Planck Institute for Evolutionary Anthropology, Leipzig, 2013.

[26] Jacob Devlin, Ming-Wei Chang, Kenton Lee, and Kristina Toutanova. Bert: Pre-training of deep bidirectional transformers for language understanding, 2018.

[27] Alexis Conneau, Kartikay Khandelwal, Naman Goyal, Vishrav Chaudhary, Guillaume Wenzek, Francisco Guzmán, Edouard Grave, Myle Ott, Luke Zettlemoyer, and Veselin Stoyanov. Unsupervised cross-lingual representation learning at scale. In *Proceedings of the 58th Annual Meeting of the Association for Computational Linguistics*, pages 8440–8451, Online, 2020. Association for Computational Linguistics.

[28] Jacek Pyżalski. From cyberbullying to electronic aggression: typology of the phenomenon. *Emotional and Behavioural Difficulties*, 17(3-4):305–317, 2012.

[29] Sara Owsley Sood, Elizabeth F. Churchill, and Judd Antin. Automatic identification of personal insults on social news sites. *J. Am. Soc. Inf. Sci. Technol.*, 63(2):270–285, 2012.

[30] Amparo Cano Basave, Kang Liu, and Jun Zhao. A weakly supervised bayesian model for violence detection in social media. In *6th International Joint Conference on Natural Language Processing (IJCNLP)*, 10 2013.

[31] Taisei Nitta, Fumito Masui, Michal Ptaszynski, Yasutomo Kimura, Rafal Rzepka, and Kenji Araki. Detecting cyberbullying entries on informal school websites based on category relevance maximization. In *Proceedings of the Sixth International Joint Conference on Natural Language Processing*, pages 579–586, Nagoya, Japan, 2013. Asian Federation of Natural Language Processing.

[32] Peter D. Turney and Michael L. Littman. Unsupervised learning of semantic orientation from a hundred-billion-word corpus. *CoRR*, cs.LG/0212012, 2002.

[33] Suzuha Hatakeyama, Fumito Masui, Michal Ptaszynski, and Kazuhide Yamamoto. Statistical analysis of automatic seed word acquisition to improve harmful expression extraction in cyberbullying detection. *International Journal of Engineering and Technology Innovation*, 6(2):165–172, 2016.

[34] "Michal Ptaszynski, Fumito Masui", Yasutomo Kimura, Rafal Rzepka, and Kenji Araki. Brute force works best against bullying. In *Proceedings of the 2015 International Conference on Constraints and Preferences for Configuration and Recommendation and Intelligent Techniques for Web Personalization - Volume 1440*, CPCR+ITWP'15, page 28–29, Aachen, DEU, 2015. CEUR-WS.org.

[35] Michal Ptaszynski, Juuso Kalevi Kristian Eronen, and Fumito Masui. Learning deep on cyberbullying is always better than brute force. In *LaCATODA 2017 CEUR Workshop Proceedings*, page 3–10, 2017.

[36] Juuso Eronen, Michal Ptaszynski, Fumito Masui, Aleksander Smywiński-Pohl, Gniewosz Leliwa, and Michal Wroczynski. Improving classifier training efficiency for automatic cyberbullying detection with feature density. *Information Processing & Management*, 58(5):102616, 2021.

[37] Sweta Agrawal and Amit Awekar. Deep learning for detecting cyberbullying across multiple social media platforms. *CoRR*, abs/1801.06482, 2018.

[38] Marzieh Mozafari, Reza Farahbakhsh, and Noël Crespi. A bert-based transfer learning approach for hate speech detection in online social media. In Hocine Cherifi, Sabrina Gaito, José Fernendo Mendes, Esteban Moro, and Luis Mateus Rocha, editors, *Complex Networks and Their Applications VIII*, pages 928–940, Cham, 2020. Springer International Publishing.

[39] Maral Dadvar and Kai Eckert. Cyberbullying detection in social networks using deep learning based models. In *International Conference on Big Data Analytics and Knowledge Discovery*, pages 245–255. Springer, 2020.

[40] J. Yadav, D. Kumar, and D. Chauhan. Cyberbullying detection using pre-trained bert model. In *2020 International Conference on Electronics and Sustainable Communication Systems (ICESC)*, pages 1096–1100, 2020.

[41] Endang Wahyu Pamungkas, Valerio Basile, and Viviana Patti. Towards multidomain and multilingual abusive language detection: a survey. *Personal and Ubiquitous Computing*, pages 1–27, 2021.

[42] Håkan Ringbom. *Cross-linguistic Similarity in Foreign Language Learning*. Multilingual Matters, 2006.

[43] Robert Bley-Vroman. The evolving context of the fundamental difference hypothesis. *Studies in Second Language Acquisition*, 31(2):175–198, 2009.

[44] Ryan Cotterell, Sabrina J. Mielke, Jason Eisner, and Brian Roark. Are all languages equally hard to language-model? In *Proceedings of the 2018 Conference of the North American Chapter of the Association for Computational Linguistics: Human Language Technologies, Volume 2 (Short Papers)*, pages 536–541, New Orleans, Louisiana, 2018. Association for Computational Linguistics.

[45] Palakorn Achananuparp, Xiaohua Hu, and Xiajiong Shen. The evaluation of sentence similarity measures. In *Proceedings of the International Conference on Data Warehousing and Knowledge Discovery*, volume 5182, pages 305–316, 09 2008.

[46] Aminul Islam and Diana Inkpen. Semantic text similarity using corpus-based word similarity and string similarity. *ACM Trans. Knowl. Discov. Data*, 2(2), 2008.

[47] Edward P. Stabler and Edward L. Keenan. Structural similarity within and among languages. *Theoretical Computer Science*, 293(2):345–363, 2003. Algebraic Methods in Language Processing.

[48] Xilun Chen, Ahmed Hassan Awadallah, Hany Hassan, Wei Wang, and Claire Cardie. Multi-source cross-lingual model transfer: Learning what to share. In *Proceedings of the 57th Annual Meeting of the Association for Computational Linguistics*, pages 3098–3112, Florence, Italy, July 2019. Association for Computational Linguistics.

[49] Quynh Do and Judith Gaspers. Cross-lingual transfer learning with data selection for large-scale spoken language understanding. In *Proceedings of the 2019 Conference on Empirical Methods in Natural Language Processing and the 9th International Joint Conference on Natural Language Processing (EMNLP-IJCNLP)*, pages 1455–1460, Hong Kong, China, 2019. Association for Computational Linguistics.

[50] Niels van der Heijden, Helen Yannakoudakis, Pushkar Mishra, and Ekaterina Shutova. Multilingual and cross-lingual document classification: A meta-learning approach. In *Proceedings of the 16th Conference of the European Chapter of the Association for Computational Linguistics: Main Volume*, pages 1966–1976, Online, April 2021. Association for Computational Linguistics.

[51] Farhad Nooralahzadeh, Giannis Bekoulis, Johannes Bjerva, and Isabelle Augenstein. Zero-shot cross-lingual transfer with meta learning. In *Proceedings of the 2020 Conference on Empirical Methods in Natural Language Processing (EMNLP)*, pages 4547–4562, Online, November 2020. Association for Computational Linguistics.

[52] Sean MacAvaney, Hao-Ren Yao, Eugene Yang, Katina Russell, Nazli Goharian, and Ophir Frieder. Hate speech detection: Challenges and solutions. *PLOS ONE*, 14(8):1–16, 08 2019.

[53] Kelly Reynolds, April Edwards, and Lynne Edwards. Using machine learning to detect cyberbullying. *Proceedings - 10th International Conference on Machine Learning and Applications, ICMLA 2011*, 2, 12 2011.

[54] Michał Ptaszynski, Gniewosz Leliwa, Mateusz Piech, and Aleksander Smywiński-Pohl. Cyberbullying detection – technical report 2/2018, department of computer science agh, university of science and technology, 2018.

[55] Michael Wiegand, Melanie Siegel, and Josef Ruppenhofer. Overview of the germeval 2018 shared task on the identification of offensive language. In *GermEval 2018 Shared Task on the Identification of Offensive Language*, 09 2018.

[56] Gudbjartur Ingi Sigurbergsson and Leon Derczynski. Offensive language and hate speech detection for Danish. In *Proceedings of the 12th Language Resources and Evaluation Conference*, pages 3498–3508, Marseille, France, May 2020. European Language Resources Association.

[57] Marcos Zampieri, Shervin Malmasi, Preslav Nakov, Sara Rosenthal, Noura Farra, and Ritesh Kumar. Predicting the type and target of offensive posts in social media. In *Proceedings of the 2019 Conference of the North American Chapter of the Association for Computational Linguistics: Human Language Technologies, Volume 1 (Long and Short Papers)*, pages 1415–1420, Minneapolis, Minnesota, June 2019. Association for Computational Linguistics.

[58] Michal Ptaszynski, Agata Pieciukiewicz, and Paweł Dybała. Results of the poleval 2019 shared task 6: First dataset and open shared task for automatic cyberbullying detection in polish twitter. In *PolEval 2019 Workshop*, pages 89–110, 2019.

[59] Michal Ptaszynski, Pawel Lempa, Fumito Masui, Yasutomo Kimura, Rafal Rzepka, Kenji Araki, Michal Wroczynski, and Gniewosz Leliwa. Brute - force sentence pattern extortion from harmful messages for cyberbullying detection. *Journal of the Association for Information Systems*, 20:8, 2019.

[60] Sergey Smetanin. Toxic comments detection in russian. In *Computational Linguistics and Intellectual Technologies: Proceedings of the International Conference "Dialogue 2020"*, 06 2020.

[61] Sameer Hinduja and Justin W Patchin. *Bullying beyond the schoolyard: Preventing and responding to cyberbullying*. Corwin Press, 2014.

[62] Masaki Arata. Study on change of detection accuracy over time in cyberbullying detection. Master's thesis, Kitami Institute of Technology, Department of Computer Science, 2019.

[63] Ippei Takenaka, Moeko Ochiai, and Yutaka Matsui. The situation of occupational stress and related factors of harmful information countermeasure workers (in japanese). *Social psychology research (Japanese Society of Social Psychology)*, 33(3):135–148, 2018.

[64] Rishi Bommasani, Drew A. Hudson, Ehsan Adeli, Russ Altman, Simran Arora, Sydney von Arx, Michael S. Bernstein, Jeannette Bohg, Antoine Bosselut, Emma Brunskill, Erik Brynjolfsson, Shyamal Buch, Dallas Card, Rodrigo Castellon, Niladri Chatterji, Annie Chen, Kathleen Creel, Jared Quincy Davis, Dora Demszky, Chris Donahue, Moussa Doumbouya, Esin Durmus, Stefano Ermon, John Etchemendy, Kawin Ethayarajh, Li Fei-Fei, Chelsea Finn, Trevor Gale, Lauren Gillespie, Karan Goel, Noah Goodman, Shelby Grossman, Neel Guha, Tatsunori Hashimoto, Peter Henderson, John Hewitt, Daniel E. Ho, Jenny Hong, Kyle Hsu, Jing Huang, Thomas Icard, Saahil Jain, Dan Jurafsky, Pratyusha Kalluri, Siddharth Karamcheti, Geoff Keeling, Fereshte Khani, Omar Khattab, Pang Wei Koh, Mark Krass, Ranjay Krishna, Rohith Kuditipudi, Ananya Kumar, Faisal Ladhak, Mina Lee, Tony Lee, Jure Leskovec, Isabelle Levent, Xiang Lisa Li, Xuechen Li, Tengyu Ma, Ali Malik, Christopher D. Manning, Suvir Mirchandani, Eric Mitchell, Zanele Munyikwa, Suraj Nair, Avanika Narayan, Deepak Narayanan, Ben Newman, Allen Nie, Juan Carlos Niebles, Hamed Nilforoshan, Julian Nyarko, Giray Ogut, Laurel Orr, Isabel Papadimitriou, Joon Sung Park, Chris Piech, Eva Portelance, Christopher Potts, Aditi Raghunathan, Rob Reich, Hongyu Ren, Frieda Rong, Yusuf Roohani, Camilo Ruiz, Jack Ryan, Christopher Ré, Dorsa Sadigh, Shiori Sagawa, Keshav Santhanam, Andy Shih, Krishnan Srinivasan, Alex Tamkin, Rohan Taori, Armin W. Thomas, Florian Tramèr, Rose E. Wang, William Wang, Bohan Wu, Jiajun Wu, Yuhuai Wu, Sang Michael Xie, Michihiro Yasunaga, Jiaxuan You, Matei Zaharia, Michael Zhang, Tianyi Zhang, Xikun Zhang, Yuhui Zhang, Lucia Zheng, Kaitlyn Zhou, and Percy Liang. On the opportunities and risks of foundation models, 2021.

[65] Ashish Vaswani, Noam Shazeer, Niki Parmar, Jakob Uszkoreit, Llion Jones, Aidan N Gomez, Łukasz Kaiser, and Illia Polosukhin. Attention is all you need. *Advances in neural information processing systems*, 30, 2017.

[66] Karthikeyan K, Zihan Wang, Stephen Mayhew, and Dan Roth. Cross-lingual ability of multilingual bert: An empirical study. In *International Conference on Learning Representations*, 2020.

[67] Shijie Wu and Mark Dredze. Beto, bentz, becas: The surprising cross-lingual effectiveness of BERT. In *Proceedings of the 2019 Conference on Empirical Methods in Natural Language Processing and the 9th International Joint Conference on Natural Language Processing (EMNLP-IJCNLP)*, pages 833–844, Hong Kong, China, November 2019. Association for Computational Linguistics.

[68] Telmo Pires, Eva Schlinger, and Dan Garrette. How multilingual is multilingual bert? In *ACL*, 2019.

[69] Yinhan Liu, Myle Ott, Naman Goyal, Jingfei Du, Mandar Joshi, Danqi Chen, Omer Levy, Mike Lewis, Luke Zettlemoyer, and Veselin Stoyanov. Roberta: A robustly optimized bert pretraining approach. *arXiv preprint arXiv:1907.11692*, 2019.

[70] Alexis Conneau and Guillaume Lample. Cross-lingual language model pretraining. *Advances in Neural Information Processing Systems*, 32:7059–7069, 2019.

[71] Cecil Brown, Eric Holman, and Søren Wichmann. Sound correspondences in the world's languages. *Language*, 89:4–29, 03 2013.

[72] Charlotte Gooskens. The contribution of linguistic factors to the intelligibility of closely related languages. *Journal of Multilingual and multicultural development*, 28(6):445–467, 2007.

[73] Junjie Hu, Sebastian Ruder, Aditya Siddhant, Graham Neubig, Orhan Firat, and Melvin Johnson. Xtreme: A massively multilingual multi-task benchmark for evaluating cross-lingual generalisation. In *International Conference on Machine Learning*, pages 4411–4421. PMLR, 2020.

[74] Yaobo Liang, Nan Duan, Yeyun Gong, Ning Wu, Fenfei Guo, Weizhen Qi, Ming Gong, Linjun Shou, Daxin Jiang, Guihong Cao, et al. Xglue: A new benchmark datasetfor cross-lingual pre-training, understanding and generation. In *Proceedings of the 2020 Conference on Empirical Methods in Natural Language Processing (EMNLP)*, pages 6008–6018, 2020.

[75] Goran Glavaš, Mladen Karan, and Ivan Vulić. XHate-999: Analyzing and detecting abusive language across domains and languages. In *Proceedings of the 28th International Conference on Computational Linguistics*, pages 6350–6365, Barcelona, Spain (Online), December 2020. International Committee on Computational Linguistics.

[76] Qiang Chen, Wenjie Li, Yu Lei, Xule Liu, and Yanxiang He. Learning to adapt credible knowledge in cross-lingual sentiment analysis. In *Proceedings of the 53rd Annual Meeting of the Association for Computational Linguistics and the 7th International Joint Conference on Natural Language Processing (Volume 1: Long Papers)*, pages 419–429, 2015.

[77] Rouzbeh Ghasemi, Seyed Arad Ashrafi Asli, and Saeedeh Momtazi. Deep persian sentiment analysis: Cross-lingual training for low-resource languages. *Journal of Information Science*, page 0165551520962781, 2020.

[78] Long Duong, Trevor Cohn, Steven Bird, and Paul Cook. Low resource dependency parsing: Cross-lingual parameter sharing in a neural network parser. In *Proceedings of the 53rd annual meeting of the Association for Computational Linguistics and the 7th international joint conference on natural language processing (volume 2: short papers)*, pages 845–850, 2015.

[79] Yuxuan Wang, Wanxiang Che, Jiang Guo, Yijia Liu, and Ting Liu. Cross-lingual bert transformation for zero-shot dependency parsing. In *Proceedings of the 2019 Conference on Empirical Methods in Natural Language Processing and the 9th International Joint Conference on Natural Language Processing (EMNLP-IJCNLP)*, pages 5721–5727, 2019.

[80] M Saiful Bari, Shafiq Joty, and Prathyusha Jwalapuram. Zero-resource cross-lingual named entity recognition. *Proceedings of the AAAI Conference on Artificial Intelligence*, 34(05):7415–7423, Apr. 2020.

[81] Yunsu Kim, Yingbo Gao, and Hermann Ney. Effective cross-lingual transfer of neural machine translation models without shared vocabularies. In *Proceedings of the 57th Annual Meeting of the Association for Computational Linguistics*, pages 1246–1257, Florence, Italy, 2019. Association for Computational Linguistics.

[82] Raj Dabre, Chenhui Chu, and Anoop Kunchukuttan. A survey of multilingual neural machine translation. *ACM Comput. Surv.*, 53(5), September 2020.